\relax
\documentclass[letterpaper]{llncs} 
\usepackage{times}  
\usepackage{helvet} 
\usepackage{courier}  
\usepackage[hyphens]{url}  
\usepackage{graphicx} 
\usepackage{graphicx}  
\usepackage{booktabs}
\usepackage{subcaption}
\usepackage{color}
\definecolor{darkgreen}{rgb}{0,0.5,0}
\definecolor{orange}{rgb}{1,0.5,0}
\definecolor{teal}{rgb}{0,0.5,0.5}
\definecolor{darkpurple}{rgb}{0.5, 0, 0.5}

\newcommand{\squishlist}{
 \begin{list}{$\bullet$}
  { \setlength{\itemsep}{0pt}
     \setlength{\parsep}{3pt}
     \setlength{\topsep}{3pt}
     \setlength{\partopsep}{0pt}
     \setlength{\leftmargin}{1.5em}
     \setlength{\labelwidth}{1em}
     \setlength{\labelsep}{0.5em} } }
\newcommand{\squishend}{
  \end{list}  }

\title{Extracting Structured Data from Physician-Patient Conversations\\ By Predicting Noteworthy Utterances}

\author{Kundan Krishna\inst{1} \and
Amy Pavel\inst{1} \and
Benjamin Schloss\inst{2} \and
Jeffrey P. Bigham\inst{1} \and
Zachary C. Lipton\inst{1}}

\institute{Carnegie Mellon University \\
\email{\{kundank,apavel,jbigham,zlipton\}@andrew.cmu.edu}
\and
Abridge AI Inc \\
\email{bschloss@abridge.ai}
}

\begin{document}    

\maketitle

\begin{abstract}

Despite diverse efforts to mine various modalities of medical data,
the conversations between physicians and patients at the time of care
remain an untapped source of insights.
In this paper, we leverage this data
to extract structured information 
that might assist physicians with post-visit 
documentation in electronic health records, 
potentially lightening the clerical burden.
In this exploratory study, we describe a new dataset consisting 
of conversation transcripts, post-visit summaries, 
corresponding supporting evidence (in the transcript), and structured labels. 
We focus on the tasks of recognizing relevant diagnoses and
abnormalities in the review of organ systems (RoS). 
One methodological challenge is that the conversations are long (around 1500 words),
making it difficult for modern deep-learning models to use them as input.
To address this challenge, we extract \emph{noteworthy} utterances---parts 
of the conversation likely to be cited as evidence supporting some summary sentence. 
We find that by first filtering for (predicted)
noteworthy utterances,
we can significantly boost predictive performance 
for recognizing both diagnoses and RoS abnormalities.

\end{abstract}

\section{Introduction}

Medical institutions collect vast amounts of patient data
in Electronic Health Records (EHRs),
including family history, past surgeries, medications and more.
Such EHR data helps physicians recall past visits, 
assess the trajectory of a patient's condition over time, 
and access crucial information (e.g., drug allergies) in emergency scenarios.
However, entering data in the EHR can be a 
tedious and time consuming for physicians. 
For every hour of visiting patients,
physicians spend around 45 minutes on EHR documentation~\cite{sinsky2016allocation},
and often need to complete documentation outside of work hours,
a significant contributor to burnout~\cite{gardner2018physician}.  
Physicians spend much of the EHR documentation time 
recalling and manually entering information 
discussed with the patient (e.g., reported symptoms).
While transcribing physician-patient discussions could aid EHR documentation, 
such conversations are long
(10 minutes / 1500 words in our dataset) 
and difficult to read due to redundancies 
and disfluencies typical of conversation.

To mitigate the burden of EHR documentation, 
we leverage transcribed physician-patient conversations
to automatically extract structured data.
As an initial investigation, 
we explore two prediction tasks 
using the physician-patient conversation as input: 
relevant diagnosis prediction, 
and organ system abnormality prediction. 
In the first task, we extract the set of diagnosis 
mentioned in the conversations
that are relevant to the chief complaint of the patient
(i.e. the purpose of the visit), 
omitting irrelevant diagnosis. 
For instance, a patient's diagnosis of hypercholestremia (high cholesterol) 
may be relevant if his visit is for hypertension 
but not relevant if the visit is for common cold.
For the second task, we recognize the organ systems 
for which the patient reported an abnormal symptom during a review.
For instance, a patient whose chief complaint is diabetes
might report fatigue (symptom) indicating 
a musculoskeletal (system) abonormality.  
Taken together, the diagnosis and symptomatic organ systems 
can provide a high-level overview of patient status 
to aid physicians in post-visit EHR documentation.

We formulate our tasks as multilabel classification 
and evaluate task performance for a medical-entity-based 
string-matching baseline, 
traditional learning approaches (e.g., logistic regression),
and state-of-the-art neural approaches (e.g., BERT). 
One challenge is that conversations are long, 
containing information irrelevant to our tasks (e.g., small talk).
A crucial finding is that a filtering-based approach 
to pre-select important parts of the conversation 
(we call them ``noteworthy'' utterances/sentences) 
before feeding them into a classification model 
significantly improves the performance of our models,
increasing micro-averaged F1 scores by 10 points for diagnosis prediction
and 5 points for RoS abnormality prediction.
We compare different ways of extracting noteworthy sentences, 
such as using a medical entity tagger and
training a model to predict such utterances,
using annotations present in our dataset. 
An oracle approach using ground truth noteworthy sentences annotated in the dataset,
boosts performance of the downstream classifiers significantly
and, remarkably, we are able to realize a significant fraction
of that gain by using our learned filters. 

We find that using sentences that are specifically noteworthy 
with respect to medical diagnoses works best for the diagnosis prediction task.
In contrast, for the RoS abnormality prediction task, 
the best performance is achieved when using sentences 
extracted by a medical entity tagger along with sentences predicted 
to be noteworthy with respect to review of systems. 

\section{Related Work}
Prior work has focused on qualitative and quantitative evaluation 
of conversations between physicians and patients,
which has been surveyed by ~\cite{roter1992quantitative}. 
Researchers have analyzed patients' questions 
to characterize their effects on the quality of interaction\cite{roter1977patient}, 
and tried to draw correlations between questioning style of physicians 
and the kind of information revealed by the patients~\cite{roter1987physicians}. 
Although research on extracting information from clinical conversations is scarce, 
there is significant work on extracting information 
from other forms of conversation such as summarizing 
email threads~\cite{murray2008summarizing} 
and decisions in meetings~\cite{wang2011summarizing}.

Compared to patient-physician conversations, 
EHR data has been heavily leveraged for a variety of tasks,
including event extraction~\cite{fries2016brundlefly},
temporal prediction~\cite{cheng2016risk}, 
and de-identification~\cite{dernoncourt2017identification}.
We point to \cite{shickel2017deep} for an overview. 
Researchers have used patient admission notes 
to predict diagnoses~\cite{li2017convolutional}. 
Using content from certain specific sections of the note 
improves performance of diagnosis extraction models 
when compared to using the entire note\cite{datla2017automated}. 
In our work too, making diagnosis predictions on 
a smaller part of conversations consisting of filtered noteworthy sentences 
leads to better model performance.
Leveraging extracted symptoms from clinical notes
using Metamap~\cite{aronson2001effective} medical ontology
improves performance on diagnosis prediction~\cite{guo2019disease}. 
This shows the usefulness of incorportaing domain knowledge for diagnosis prediction, 
which we have also leveraged for our tasks by using a medical entity tagging system.
Beyond diagnosis prediction, 
EHR data has been used to extract other information 
such as medications and lab tests~\cite{wu2015named}, 
including fine-grained information like dosage 
and frequency of medicines and severity of diseases~\cite{jagannatha2016bidirectional}.

The systems in all of this work are based on clinical notes
in the EHR, which are abundant in datasets. 
The research on information extraction from medical conversations 
is scarce likely owed in part to the paucity of datasets
containing both medical conversations and annotations. 
Creating such a dataset is difficult due to the medical expertise
that is required to annotate medical conversations with tags 
such as medical diagnoses and lab test results. 
One notable work in this area extracts symptoms from patient-physician conversations~\cite{rajkomar2019automatically}. 
Their model takes as input snippets of 5 consecutive utterances 
and predicts whether the snippet has a symptom mentioned and experienced
by the patient, using a recurrent neural network. 
In contrast, we make predictions of diagnoses and RoS abnormalities 
from an entire conversation using a variety of models
including modern techniques from deep NLP, 
and introduce an approach to aid this 
by filtering out noteworthy sentences from the conversation.



\section{Dataset}

\begin{figure*}
        \centering
        \begin{subfigure}[t]{0.3\textwidth}  
            \centering 
            \includegraphics[width=\textwidth]{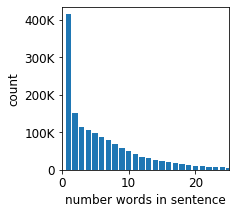}
            \caption{\centering
            Number of words per conversation sentence
            }    
            \label{fig:transcriptsent_numsentwords}
        \end{subfigure}
        \begin{subfigure}[t]{0.3\textwidth}   
            \centering 
            \includegraphics[width=\textwidth]{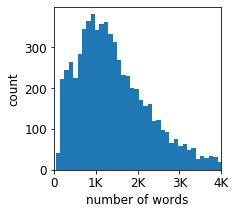}
            \caption{\centering
Histogram of number of words per conversation             }   
            \label{fig:transcript_numconvwords}
        \end{subfigure}
        \begin{subfigure}[t]{0.3\textwidth}   
            \centering 
            \includegraphics[width=\textwidth]{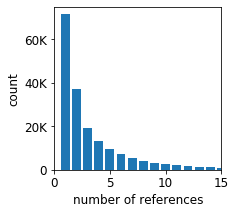}
            \caption{\centering
            Number of reference utterances from conversation used for a SOAP note entry
            }    
            \label{fig:dist_numrefs}
        \end{subfigure}
        \caption{
        Distribution of sentence lengths, number of words in physician-patient conversations of our dataset, and the  number of evidence utterances in it referred by an entry of the corresponding SOAP note.}
        \label{fig:datasetstats}
    \end{figure*}

This paper addresses a dataset of human-transcribed physician-patient conversations. 
The dataset includes 2732 cardiologist visits, 2731 family medicine visits,
989 interventional cardiologist visits, and 410 internist visits. 
Each transcript consists of timestamped utterances with speaker labels. 
A typical conversation consists of $200$-$250$ utterances. 
The median utterance is short (Figure~\ref{fig:transcriptsent_numsentwords}), 
possibly due to the high frequency of back-chanelling (e.g., ``umm-hmm'', ``okay'', etc.). 
In total, each conversation contains around 1500 words (Figure \ref{fig:transcript_numconvwords}).

In our dataset, the transcribed conversations are coupled 
with corresponding structured text summaries and summary annotations.
The structured text summaries (SOAP notes) 
are typically written by a physician to summarize a patient visit, 
and their annotations were constructed
by expert clinical annotators 
who received task-specific training.
The acronym SOAP in SOAP note stands for 
the four sections of the note: 
The (S)ubjective section contains
a subjective accounting of the patient's current symptoms,
and a history of the present illness, and miscellaneous details.
The (O)bjective section contains objective information 
such as results from lab tests,
and observations from a physical examination. 
The (A)ssessment and (P)lan sections contain the inferences 
made by the physician, including the differential diagnosis,
and the plan for treatment, including  further tests,
planned changes to the patient's medications, 
other non-pharmaceutical therapeutics, and more. 

In total, our dataset consists of 6862 datapoints 
(i.e., physician-patient conversation transcripts with corresponding annotated notes), 
which we have then divided into train and test sets
with sizes $6270$ and $592$, respectively. 
To train our models, we set aside $500$ points 
as a validation set for tuning hyperparmeters. 
The number of datapoints and the splits 
are the same for both the tasks.

In our dataset, each line in a SOAP note is classified 
into one of 12 total subsections within one of the high-level 
Subjective, Objective, Assessment, or Plan sections.
For example, subsections for the Subjective section include 
\textit{Subjective: Review Of Systems} and \textit{Subjective:Past Medical History}.
Each line in a SOAP note appears alongside 
structured categorical or numerical metadata. 
For instance, a SOAP note line about medication 
(e.g., ``Take Aspirin once a day.'') 
may be coupled with structured data 
for the medication name (e.g., ``Aspirin'') 
and the dosage (e.g., ``daily''). 
Each SOAP note line is also associated with
the lines in the transcript that were used 
as evidence by the annotator create the line and its metadata. 
Each SOAP note line with its associated metadata, i.e., SOAP note entry,
uses an average of $3.85$ transcript lines as evidence (Figure \ref{fig:dist_numrefs}).
We take subsets of information from the dataset described above 
to design datasets for the relevant diagnosis prediction 
and review of systems abnormality prediction tasks.

\subsection{Relevant Diagnosis Prediction}
Given a physician-patient conversation, 
we aim to extract the mentioned past and present diagnoses of the patient 
that are relevant to the primary reason 
for the patient's visit (called the Chief Complaint).
For each conversation, we create a list of the Chief Complaint 
and related medical problems by using categorical tags 
associated with the following subsections of the SOAP note:
\begin{enumerate}
    \item The Chief Complaint of the patient from~\textit{Subjective: Chief Complaint} the subsection of the SOAP note. 
    \item All medical problems in the \textit{Subjective: Past Medical History} subsection tagged with ``HPI'' (History of Present Illness) to signify that they are related to the Chief Complaint.
    \item The medical problem tags present in the \textit{Assessment and Plan: Assessment}
    subsection of the SOAP note. 
\end{enumerate}

We then simplified the medical problem tags by converting everything to lowercase, 
and removing elaborations given in parentheses. 
For example, we simplify ``hypertension (moderate to severe)'' to “hypertension''.
For each of the 20 most frequent tags retrieved after the previous simplifications, 
we searched among all medical problems 
and added the ones that had the original tag as a substring.
For example, ``systolic hypertension'' was merged into ``hypertension''.
After following the above procedure on the training and validation set,
we take the 15 most frequent medical problem tags 
(Table \ref{tab:diagnosesrosfreq}) 
and restrict the task to predicting 
whether each of these medical problems 
were diagnosed for a patient or not. 

\begin{table}[t]
\centering
\begin{subtable}{0.48\textwidth}
    \raggedleft
    \begin{tabular}{l c}
    \toprule
    Diagnosis & Frequency\\
    \midrule
    hypertension & 1573 \\
    diabetes & 1423 \\
    atrial fibrillation & 1335 \\
    hypercholesterolemia & 1023 \\
    heart failure & 584 \\
    myocardial infarction & 386 \\
    arthritis & 288 \\
    cardiomyopathy & 273 \\
    coronary arteriosclerosis & 257 \\
    heart disease & 240 \\
    chronic obstructive lung disease & 235 \\
    dyspnea & 228 \\
    asthma & 188 \\
    sleep apnea & 185 \\
    depression & 148 \\
    \bottomrule    
    \end{tabular}
\end{subtable}
\begin{subtable}{0.38\textwidth}
    \centering
    \begin{tabular}{l c}
    \toprule    
    System & Frequency\\
    \midrule
    cardiovascular & 2245\\
    musculoskeletal & 1924\\
    respiratory & 1401\\
    gastrointestinal & 878\\
    skin & 432\\
    head & 418\\
    neurologic & 385\\
  \bottomrule    
    \end{tabular}
\end{subtable}    
    
    \caption{
    Diagnoses and abnormal systems extracted from the train+validation split of the dataset with their number of occurrences}
    \label{tab:diagnosesrosfreq}
\end{table}

\subsection{Review of Systems (RoS) Abnormality Prediction}
Given a physician-patient conversation,
we also predict the organ systems (e.g., respiratory system) 
for which the patient predicted a symptom (e.g., trouble breathing). 
During a patient's visit, the physician conducts a Review of Systems (RoS),
where the physician reviews organ systems 
and potential associated symptoms and asks if the patient is experiencing each symptom.
In our dataset SOAP notes, the \textit{Subjective: Review of Systems} subsection 
contains annotated observations from the RoS, 
each containing a system, symptom, and result. 
For instance, a system (e.g., ``cardiovascular''), 
an associated symptom (e.g., ``chest pain or discomfort'') 
and a result based on patient feedback (e.g., ``confirms'', ``denies'').
To reduce sparsity in the data for system/symptom pairs, 
we consider only systems and whether or not 
each system contained a confirmed symptom.
We also consider only the set of 7 systems 
for which more than 5\% of patients reported abnormalities,
for prediction (Table \ref{tab:diagnosesrosfreq}).

\section{Methods}
We use a single suite of models for both tasks.

\subsubsection{Input-agnostic baseline}
We establish the best value of each metric
that can be achieved without using the input 
(i.e., an input-agnostic classifier).
The behavior of the input-agnostic classifier depends on the metric.
For example, to maximize accuracy,
the classifier predicts the majority class 
(usually negative) for all diagnoses.
On the other hand, to maximize F1 and recall, 
the classifier predicts the positive class for all diagnoses. 
To maximize AUC and precision-at-1, 
the classifier assigns probabilities to each diagnosis 
according to their prevalence rates.
For a detailed description of multilabel performance metrics, 
we point to~\cite{lipton2014optimal}.

\subsubsection{Medical-entity-matching baseline}
This baseline uses a traditional string-matching tool. 
For extracting relevant diagnoses, for each diagnosis, 
we check to see whether it is mentioned in the conversation. 
Since a diagnosis can be expressed in different ways,
e.g., ``myocardial infarction'' has the same meaning as the common term ``heart attack'',
we use a system for tagging medical terms (QuickUMLS) 
that maps strings to medical entities with a unique ID. 
For example, ``hypertension'' and ``high blood pressure'' 
are both mapped to the same ID.

For predicting RoS abnormalities, our baseline predicts 
that the person has an abnormality in a system 
if any symptom related to the system is mentioned 
in the text as detected by QuickUMLS.
The symptoms checked for each system are taken 
from the RoS tags in the dataset.
For example, the cardiovascular system has symptoms 
like ``chest pain or discomfort'' and ``palpitations, shortness of breath''.

\subsubsection{Learning based methods}
We apply the following classical models: 
Logistic Regression, Support Vector Classifier, 
Multinomial Naive Bayes, Random Forest, and Gradient Boosting. 
We use bag-of-words representation of conversations 
with unigrams and bigrams with TF-IDF transform on the features. 

We also applied state of the art neural methods on the problem. 
We classified diagnoses and RoS abnormalities as present or not 
using two BERT models with wordpiece~\cite{wu2016google} tokenization--- one 
generic, pretrained BERT model,
and one pretrained BERT model that is finetuned 
on clinical text~\cite{alsentzer-etal-2019-publicly}. 
Each of our BERT models are 12-layered with a hidden size of 768. 
The final hidden state of the [CLS] token is taken
as the fixed-dimensional pooled representation of the input sequence. 
This is fed into a linear layer with sigmoid activation and output size 
equal to the number of prediction classes
($15$ for diagnosis prediction and $7$ for the RoS abnormality prediction), 
thus giving us the probability for each class. 
Since the pretrained BERT models 
do not support a sequence length of more than 512 tokens, 
we break up individual conversations into chunks of 512 tokens, 
pass the chunks independently through BERT,
and mean-pool their [CLS] representations.
Due to memory constraints,
we only feed the first 2040 tokens of a conversation into the model.

\subsection{Hybrid models}  
The long length of the input sequence 
makes the task difficult for the neural models. 
We tried a variety of strategies to pre-filter the contents of the conversation 
so that we only feed in sentences that are more relevant to the task. 
We call such sentences \textit{noteworthy}. 
We have 3 ways for deciding if a sentence is noteworthy, 
which lead to 3 kinds of noteworthy sentences.
\begin{itemize}
    \item \textbf{UMLS-noteworthy:} 
    We designate a sentence as noteworthy if the QuickUMLS medical tagger 
    finds an entity relevant to the task (e.g., a diagnosis or symptom) 
    as defined in the medical-entity-matching baseline.
    \item \textbf{All-noteworthy:} We deem a sentence in the conversation noteworthy 
    if it was used as evidence for any line in the annotated SOAP note. 
    We train a classifier to predict the noteworthy sentences given a conversation.
    \item \textbf{Diagnosis/RoS-noteworthy:} 
    Here, only those sentences  that were used as evidence 
    for an entry containing the ground truth tags(diagnosis/RoS abnormality) 
    that we are trying to predict are deemed noteworthy.
\end{itemize}

In addition to trying out these individual filtering strategies,
we also try their combinations as we shall discuss in the following section.

\begin{table*}[h!]
    \centering
\begin{tabular}{lcccccc}
\toprule
\textbf{Model} & \textbf{Accuracy} & \textbf{M-AUC} & \textbf{M-F1} & \textbf{m-AUC} & \textbf{m-F1} & \textbf{Precision-at-1} \\
\midrule
Input agnostic baseline & 0.9189 & 0.5000 & 0.1414 & 0.7434 & 0.3109 & 0.2027 \\ 
UMLS Medical Entity Matching & 0.9122 & 0.8147 & 0.5121 & 0.8420 & 0.5833 & 0.5034 \\
Logistic Regression & 0.9417 & 0.8930 & 0.2510 & 0.9317 & 0.5004 & 0.6064 \\
LinearSVC & 0.9395 & 0.8959 & 0.2113 & 0.9354 & 0.4603 & 0.6199 \\
Multinomial NaiveBayes & 0.9269 & 0.7171 & 0.0615 & 0.8296 & 0.1938 & 0.4848 \\
Random Forest & 0.9212 & 0.8868 & 0.0155 & 0.8795 & 0.0541 & 0.5304 \\
Gradient Boosting Classifier & 0.9467 & 0.9181 & 0.5024 & 0.9447 & 0.6514 & 0.5861 \\
BERT & 0.9452 & 0.8953 & 0.4413 & 0.9365 & 0.6009 & 0.6199 \\
CLINICALBERT (CBERT) & 0.9476 & 0.9040 & 0.4573 & 0.9413 & 0.6029 & 0.6300 \\
AN+CBERT & 0.9511 & 0.9222 & 0.4853 & 0.9532 & 0.6561 & 0.6470 \\
DN+CBERT & 0.9551 & \textbf{0.9342} & \textbf{0.5655} & \textbf{0.9616} & 0.7029 & \textbf{0.6621} \\
UMLS+CBERT & 0.9519 & 0.8615 & 0.5238 & 0.9290 & 0.6834 & 0.6030 \\
UMLS-AN-CBERT & 0.9541 & 0.9261 & 0.5317 & 0.9588 & 0.6803 & \textbf{0.6621} \\
UMLS-DN-CBERT & 0.9510 & 0.9359 & 0.5210 & 0.9593 & 0.6641 & 0.6368 \\
UMLS-F2K-AN+CBERT & \textbf{0.9554} & 0.9188 & 0.5599 & 0.9567 & \textbf{0.7139} & 0.6487 \\
UMLS+F2K-DN+CBERT & 0.9535 & 0.9354 & 0.5301 & 0.9610 & 0.6911 & 0.6486 \\
\textbf{(Oracle)} AN+CBERT & 0.9509 & 0.9418 & 0.5500 & 0.9588 & 0.6789 & 0.6250 \\
\textbf{(Oracle)} DN+CBERT & 0.9767 & 0.9771 & 0.7419 & 0.9838 & 0.8456 & 0.7162 \\
\bottomrule
\end{tabular}
    \caption{Aggregate results for the medical diagnosis prediction task. AN: predicted noteworthy utterances, DN: utterances predicted to be noteworthy specifically concerning a summary passage discussing diagnoses, F2K: UMLS-extracted noteworthy utterances with added top predicted AN/DN utterances to get K total utterances, M-: macro average, m-: micro average}
    \label{tab:diagnosisaggregate}
\end{table*}
\begin{table*}[]
    \centering
        \begin{tabular}{lcccccc}
        \toprule
        \textbf{Model} & \textbf{Accuracy} & \textbf{M-AUC} & \textbf{M-F1} & \textbf{m-AUC} & \textbf{m-F1} & \textbf{Precision-at-1} \\
        \midrule
            Input agnostic baseline & 0.8677 & 0.5000 & 0.2235 & 0.7024 & 0.3453 & 0.3040 \\  
            UMLS Medical Entity Matching & 0.4532 & 0.7074 & 0.2797 & 0.7454 & 0.3079 & 0.3226 \\
            Logistic Regression & 0.8819 & 0.8050 & 0.2102 & 0.8496 & 0.3506 & 0.3952 \\
            LinearSVC & 0.8798 & 0.8093 & 0.1623 & 0.8516 & 0.3025 & 0.3986 \\
            Multinomial NaiveBayes & 0.8687 & 0.6183 & 0.0369 & 0.7383 & 0.0653 & 0.3818 \\
            Gradient Boosting Classifier & 0.8740 & 0.7949 & 0.2500 & 0.8405 & 0.3324 & 0.4020 \\
            Random Forest & 0.8677 & 0.7210 & 0.0000 & 0.7670 & 0.0000 & 0.3412 \\
            BERT & 0.8818 & 0.8240 & 0.3304 & 0.8620 & 0.4275 & 0.3986 \\
            CLINICALBERT (CBERT) & 0.8784 & 0.8305 & 0.3878 & 0.8667 & 0.4857 & 0.4003 \\
            AN+CBERT & \textbf{0.8837} & 0.8491 & 0.3560 & 0.8801 & 0.4761 & 0.4274 \\
            RN+CBERT & 0.8861 & 0.8391 & 0.3720 & 0.8788 & 0.4925 & 0.4054 \\
            UMLS+CBERT & 0.8769 & 0.8036 & 0.3421 & 0.8464 & 0.4457 & 0.3902 \\
            UMLS+AN+CBERT & 0.8868 & 0.8252 & 0.3039 & 0.8626 & 0.4515 & 0.4139 \\
            UMLS+RN+CBERT & 0.8810 & 0.8390 & 0.3122 & 0.8745 & 0.4152 & 0.3902 \\
            UMLS+F2K-AN+CBERT & 0.8834 & 0.8169 & 0.2385 & 0.8585 & 0.3894 & 0.4189 \\
            UMLS+F2K-RN+CBERT & 0.8827 & \textbf{0.8595} & \textbf{0.3987} & \textbf{0.8895} & \textbf{0.5308} & \textbf{0.4291} \\            
            \textbf{(Oracle)} AN+CBERT & 0.8846 & 0.8535 & 0.3662 & 0.8841 & 0.5062 & 0.4375 \\
            \textbf{(Oracle)} RN+CBERT & 0.9454 & 0.9595 & 0.7235 & 0.9703 & 0.7847 & 0.4966 \\
        \bottomrule
        \end{tabular}
    \caption{Aggregate results for the RoS abnormality prediction task. AN: predicted noteworthy utterances, RN: utterances predicted to be noteworthy specifically concerning a summary passage discussing review of systems, F2K: UMLS-extracted noteworthy utterances with added top predicted AN/RN utterances to get K total utterances, M-: macro average, m-: micro average}
    \label{tab:rosaggregate}
\end{table*}
\begin{table*}[]
    \centering
\begin{tabular}{lccccccc}
\toprule
\textbf{Disease} & \textbf{Prevalence rate} & \textbf{Precision} & \textbf{Recall} & \textbf{F1} & \textbf{Accuracy} & \textbf{AUC} & \textbf{CP@1} \\
\midrule
atrial fibrillation & 0.2568 & 0.8667 & 0.9408 & 0.9022 & 0.9476 & 0.9773 & 0.3597 \\
hypertension & 0.2027 & 0.6667 & 0.4833 & 0.5604 & 0.8463 & 0.8817 & 0.0995 \\
diabetes & 0.1959 & 0.8411 & 0.7759 & 0.8072 & 0.9274 & 0.9586 & 0.1837 \\
hypercholesterolemia & 0.1216 & 0.5694 & 0.5694 & 0.5694 & 0.8953 & 0.9246 & 0.0740 \\
heart failure & 0.1014 & 0.8049 & 0.5500 & 0.6535 & 0.9409 & 0.9692 & 0.0638 \\
myocardial infarction & 0.0861 & 0.8571 & 0.8235 & 0.8400 & 0.9730 & 0.9857 & 0.0995 \\
coronary arteriosclerosis & 0.0372 & 0.3846 & 0.2273 & 0.2857 & 0.9578 & 0.8307 & 0.0051 \\
chronic obstr. lung disease & 0.0372 & 0.7391 & 0.7727 & 0.7556 & 0.9814 & 0.9665 & 0.0281 \\
dyspnea & 0.0304 & 0.5000 & 0.0556 & 0.1000 & 0.9696 & 0.9068 & 0.0077 \\
depression & 0.0304 & 0.6471 & 0.6111 & 0.6286 & 0.9780 & 0.9555 & 0.0230 \\
asthma & 0.0287 & 0.8462 & 0.6471 & 0.7333 & 0.9865 & 0.9951 & 0.0230 \\
cardiomyopathy & 0.0236 & 0.7143 & 0.7143 & 0.7143 & 0.9865 & 0.9779 & 0.0128 \\
heart disease & 0.0236 & 0.0000 & 0.0000 & 0.0000 & 0.9764 & 0.7058 & 0.0026 \\
arthritis & 0.0220 & 0.3636 & 0.3077 & 0.3333 & 0.9730 & 0.9843 & 0.0128 \\
sleep apnea & 0.0186 & 0.6667 & 0.5455 & 0.6000 & 0.9865 & 0.9937 & 0.0051 \\
\bottomrule
\end{tabular}
    \caption{
    Performance of our best diagnosis prediction model (DN+CBERT) at predicting individual diagnoses. CP@1: contribution to precision-at-1, the fraction of times a disease was a correct top prediction 
    }
    \label{tab:diagnosiswiseresults}
\end{table*}
\setlength\tabcolsep{3.8pt}

\begin{table*}[]
    \centering
    \begin{tabular}{lccccccc}
    \toprule
    \textbf{System} & \textbf{Prevalence rate} & \textbf{Precision} & \textbf{Recall} & \textbf{F1} & \textbf{Accuracy} & \textbf{AUC} & \textbf{CP@1} \\
    \midrule
    cardiovascular & 0.3041 & 0.5867 & 0.7333 & 0.6519 & 0.7618 & 0.8475 & 0.5079 \\
    musculoskeletal & 0.2010 & 0.5893 & 0.5546 & 0.5714 & 0.8328 & 0.8579 & 0.2402 \\
    respiratory & 0.1571 & 0.5231 & 0.3656 & 0.4304 & 0.8480 & 0.8639 & 0.1063 \\
    gastrointestinal & 0.0845 & 0.5217 & 0.4800 & 0.5000 & 0.9189 & 0.8636 & 0.0669 \\
    head & 0.0828 & 0.4412 & 0.3061 & 0.3614 & 0.9105 & 0.9252 & 0.0591 \\
    neurologic & 0.0574 & 0.0000 & 0.0000 & 0.0000 & 0.9426 & 0.7864 & 0.0000 \\
    skin & 0.0389 & 0.6667 & 0.1739 & 0.2759 & 0.9645 & 0.8719 & 0.0197 \\
    \bottomrule
    \end{tabular}
    \caption{
    Performance of our best RoS abnormality prediction model (UMLS+F2K-RN+CBERT) at predicting abnormalities in each system. CP@1: contribution to precision-at-1, the fraction of times an RoS abnormality was a correct top prediction
    }
    \label{tab:systemwise}
\end{table*}

\section{Results and Discussion}
\subsection{Metrics}

We evaluate the performance of models using the following metrics: 
accuracy, area under the receiver-operator characteristics (AUC), F1 score, and precision-at-1.
Because this is a $15$-label multilabel classification task 
reporting aggregate scores across labels requires some care.
For both F1 and AUC, we aggregate scores 
using both micro- and macro-averaging~\cite{van2013macro}
following the metrics for multilabel diagnosis prediction in \cite{lipton2015learning}.
Macro-averaging averages scores calculated separately on each label, 
while micro-averaging pools predictions across labels 
before calculating a single metric. 
We also compute precision-at-1 to capture the percentage of times
that each model's most confident prediction is correct
(i.e., the frequency with which the most confidently predicted diagnosis actually applies).

\subsection{Results}
We evaluated the performance of all models aggregated across classes 
on the tasks of relevant diagnosis prediction
(Table \ref{tab:diagnosisaggregate}) 
and RoS abnormality prediction (Table \ref{tab:rosaggregate}). 
Predicting RoS abnormality proves to be a more difficult task 
than predicting relevant diagnoses as reflected 
by the lower values achieved on all metrics.
We hypothesize that this is because of the variety of symptoms 
that can be checked for each system. 
The cardiovascular system has $152$ symptoms in our dataset,
including `pain in the ribs',  `palpitations', 
`increased heart rate' and `chest ache'.
A learning-based model would must learn to associate
all of these symptoms with the cardiovascular system 
in addition to recognizing whether or not 
any given patient experiences the symptom. 

For diagnosis prediction, medical-entity-matching baseline 
achieves better F1 scores than many of the classical models (Table~\ref{tab:diagnosisaggregate}). 
The high recall and low precision together demonstrate 
that if a  diagnosis has been made for the patient, 
the diagnosis is often directly mentioned in the conversation but the converse is not true.
Among the BERT-based models, we see 
a modest improvement in F1 and precision-at-1
when using ClinicalBERT instead of the common BERT.
Using predicted noteworthy sentences from the transcript 
instead of all of the transcript generally 
yielded performance gains. 
For diagnosis prediction, models 
that only ingest predicted diagnosis-noteworthy sentences 
rather than all-noteworthy sentences 
perform the best for a majority of the metrics. 
For RoS abnormality prediction, 
the trend reverses and using predicted RoS-noteworthy sentences 
performs worse than using predicted all-noteworthy sentences from the transcript. 
If we train on oracle noteworthy sentences, 
we achieve a  precision-at-1 of $0.72$ 
for diagnosis prediction and $0.50$ for RoS abnormality prediction. 
Note that the maximum achievable precision-at-1 
on the diagnosis prediction task is $0.7584$  
and for the RoS abnormality prediction task it is $0.5811$, 
because the patients do not always have one of the diagnoses 
or RoS abnormalities that we are concerned with. 

The average number of UMLS-noteworthy sentences extracted 
by QuickUMLS for diagnosis prediction and RoS abnormality prediction tasks 
is $4.42$ and $5.51$, respectively,
out of an average of $215.14$ total sentences.
We train BERT on only the UMLS-noteworthy sentences, 
on a union of UMLS-noteworthy sentences
and predicted all/task-specific noteworthy sentences,
and a FillUptoK(F2K) variant,
where we take the union but we only add the 
top-predicted all/task-specific noteworthy sentences 
until we reach a total of $K$ sentences to be fed into BERT, 
where $K$ is a hyperparameter. 
The last model achieves the best results for RoS abnormality prediction 
when we pool the UMLS-noteworthy sentences 
with the top-predicted RoS-noteworthy sentences. 
This is in contrast with results for disease prediction 
where only using predicted disease-noteworthy sentences performed best. 
This suggests that domain knowledge in the medical tagger 
may be more useful for RoS-abnormality prediction, 
because it gives an explicit listing of 
the large variety of symptoms pertaining to each organ system.

\begin{figure}[h!]
    \centering
    \includegraphics[width=0.65\textwidth]{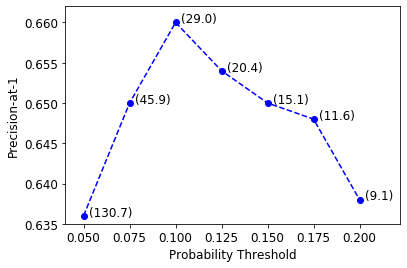}
    \caption{Precision-at-1 at different thresholds of the diagnosis-noteworthy utterance classifier. Average number of noteworthy sentences extracted in parantheses}
    \label{fig:varythreshold}
\end{figure}

On both tasks, it is possible to use 
a small fraction of the transcript 
and still get performance comparable to models that use all of it. 
For the task of diagnosis prediction, 
the UMLS-noteworthy sentences only make up 
$2.1\%$ of total sentences in a conversation on average,
but using just them with the ClinicalBERT model 
still achieves higher F1 scores than all classical ML models
which use the entire conversation.
We carried out an experiment to observe the correlation 
between number of words input into the ClinicalBERT model and the performance achieved. 
To do this, we varied the threshold probability 
for the noteworthy utterance classifier 
in the  Diagnosis-noteworthy+ClinicalBERT model.
Fewer noteworthy sentences are extracted and 
passed to ClinicalBERT as the threshold goes up. 
The performance increases with a decrease in 
the number of filtered sentences and then 
goes down (Figure~\ref{fig:varythreshold}). 
The best performance is achieved when we pass 
an average of $29$ utterances for each transcript.

\subsubsection{Performance on binary prediction tasks}
Besides calculating the aggregate performance of our models, 
we also compute the performance of our best model 
for each task at the binary prediction 
of each diagnosis/RoS abnormality 
(Table \ref{tab:diagnosiswiseresults} and Table \ref{tab:systemwise}).
We see that generally diagnoses that 
are more common are detected better. 
One exception is hypertension which has a low recall and precision
despite affecting around $20\%$ of the patients. 
The instances of hypertension that are not identified by our model 
show that it is rarely mentioned explicitly during conversation. 
Instead, it needs to be inferred by values of blood pressure readings
and phrases like ``that blood pressure seems to creep up a little bit''. 
This indirect way in which hypertension is mentioned 
possibly makes it harder to detect accurately.
In contrast, atrial fibrillation is 
usually mentioned explicitly during conversation, 
which is why even the medical-entity-matching baseline 
achieves a high recall of $0.83$ at predicting atrial fibrillation.
The model has the worst performance for predicting heart disease.
We think it is  due to a combination of low frequency 
and the generic nature of the class. 
We found that the heart disease tag is used in miscellaneous situations 
like genetic defect, weakness in heart's function, or pain related to stent placement.

We also calculate the contribution to 
precision-at-1 for each class for both tasks. 
This gives us a sense of how often a diagnosis/RoS abnormality 
becomes the model's top prediction. 
We do not want a situation 
where only the most frequent diagnoses/RoS abnormalities 
are predicted with the highest probability 
and the rarer classes are never representated among the top predictions. 
We define the contribution to precision-at-1 for a class 
as the number of times it was a correct top prediction 
divided by the total number of correct top predictions made by the model. 
We see that for both tasks, contribution to precision-at-1 
is roughly in proportion to the prevalence rate of each diagnosis 
(Table ~\ref{tab:diagnosiswiseresults} and Table~\ref{tab:systemwise}). 
This suggests that the model predicts even the rarer diagnoses 
with enough confidence for them to show up as top predictions.

\subsection{Experimental details}
The hyperparameters of each learning based model
are determined by tuning over the validation set. 
All models except the neural network based ones 
are sourced from \texttt{scikit-learn} \cite{scikit-learn}. 
The UMLS based tagging system is taken from \cite{soldaini2016quickumls}. 
The BERT-based models are trained in AllenNLP \cite{Gardner2017AllenNLP}. 
The vanilla BERT model is the bert-base-uncased model 
released by Google and the clinical BERT model 
is taken from \cite{alsentzer-etal-2019-publicly}.

The BERT models have a learning rate of $0.00002$. 
We tuned the probability threshold for predicting noteworthy sentences. 
The optimal threshold was $0.4$ for predicting all noteworthy sentences, 
$0.1$ for predicting diagnosis-related noteworthy sentences,
and $0.02$ for predicting RoS-related noteworthy sentences.
Among the FillUptoK predictors for diagnosis prediction, 
the one using AllNoteworthy sentences had $K=50$ 
and the one using diagnosis-noteworthy sentences has $K=15$.
For the FillUptoK predictors used for RoS abnormality prediction, 
the one using all-noteworthy sentences had $K=50$ 
and the predictor using RoS-noteworthy sentences had $K=20$. 

The noteworthy sentence extractors are logistic regression models trained, 
validated and tested on the same splits of the dataset as the other models. 
All models are L2-regularized with the regularization constant equal to $1$. 
The AUC scores for the classifiers extracting all, 
diagnosis-related, and RoS-related noteworthy sentences 
are $0.6959$, $0.6689$ and $0.7789$ respectively.

\section{Conclusion and Future Work}
This work is a preliminary investigation into the utility
of medical conversations for drafting SOAP notes. 
Although we have only tried predicting diagnoses and review of systems, 
there are more tasks that can be attacked using the annotations in the dataset we used
(e.g., medications, future appointments, lab tests etc). 
Our work shows that extracting noteworthy sentences 
improves the performance significantly. 
However, the performance of noteworthy sentence extractors 
leaves room for improvement, a promising direction for future work. 
Currently, we are only predicting the organ system 
that has a reported symptom and not the exact symptom 
that was reported by the patient. 
This was because the frequency of occurrence of each symptom was fairly low. 
In future work, we plan to explore models capable of performing 
better on the long tail in clinical prediction tasks.

\section{Acknowledgements}
We gratefully acknowledge support from the Center for Machine Learning and Health in a joint venture between UPMC and Carnegie Mellon University and Abridge AI, who created the dataset that we used for this research.

\bibliography{main} 
\bibliographystyle{splncs04}

\end{document}